\documentclass[10pt,twocolumn,letterpaper]{article}

\usepackage{cvpr}              

\usepackage{multirow}
\usepackage{booktabs}
\usepackage{amssymb}
\usepackage{xcolor}
\usepackage{amsmath}

\definecolor{cvprblue}{rgb}{0.21,0.49,0.74}
\usepackage[pagebackref,breaklinks,colorlinks,allcolors=cvprblue]{hyperref}

\def\paperID{4816} 
\def\confName{CVPR}
\def\confYear{2025}

\title{{\vspace{-40pt}}Structure-Aware Correspondence Learning for Relative Pose Estimation}

\author{
\quad Yihan Chen$^{1}$
\quad Wenfei Yang$^{1}$
\quad Huan Ren$^{1}$
\quad Shifeng Zhang$^{3}$
\quad Tianzhu Zhang$^{1,2}$\footnotemark[1]
\quad Feng Wu$^{1}$\\
$^1$University of Science and Technology of China \\
$^2$National Key Laboratory of Deep Space Exploration, Deep Space Exploration Laboratory \\
$^3$Sangfor Technologies\\
{\tt\small \{yihanchen, rh\_hr\_666\}@mail.ustc.edu.cn, zhangshifeng@sangfor.com.cn} \\
{\tt\small \{yangwf, tzzhang, fengwu\}@ustc.edu.cn}
}

\begin{document}
\maketitle
\renewcommand{\thefootnote}{\fnsymbol{footnote}}
\footnotetext[1]{Corresponding author.} 
\vspace{-25pt}
\begin{abstract}
Relative pose estimation provides a promising way for achieving object-agnostic pose estimation. Despite the success of existing 3D correspondence-based methods, the reliance on explicit  feature matching suffers from small overlaps in visible regions and  unreliable  feature estimation for invisible regions.  Inspired by humans' ability to  assemble two object parts that have small or no overlapping regions by considering object structure,  we propose a novel Structure-Aware Correspondence Learning method for Relative Pose Estimation, which consists of two key modules. First, a structure-aware keypoint extraction module is designed to locate a set of kepoints that can represent the structure of objects with different shapes and appearance, under the guidance of a keypoint based image reconstruction loss. Second, a structure-aware correspondence estimation module is designed to model the intra-image and inter-image relationships between keypoints to extract structure-aware features for correspondence estimation. By jointly leveraging these two modules,  the proposed method can naturally estimate 3D-3D correspondences for unseen objects without explicit feature matching for precise relative pose estimation. Experimental results on the CO3D, Objaverse and LineMOD datasets demonstrate that the proposed method significantly outperforms prior methods, i.e., with $5.7^\circ$ reduction in mean angular error on the CO3D dataset. Our code is available at \url{https://github.com/Cyhhzo02/SAC-Pose-code}.
\vspace{-5pt}
\end{abstract}    
\vspace{-10pt}

\section{Introduction}
\label{sec:intro}
\vspace{-5pt}
Object pose estimation aims to estimate the 3D translation and 3D rotation of an object from a single image. It plays a crucial role in many real-world applications such as augmented reality (AR)~\cite{azuma1997survey,marchand2015pose}, robotic manipulation~\cite{mousavian20196,wu2020grasp,wen2022you} and autonomous driving~\cite{chen2017multi,geiger2012we}, drawing increasing attention in recent years. 
Early works~\cite{xiang2017posecnn,wang2019densefusion,he2020pvn3d,he2021ffb6d} mainly focus on instance-level pose estimation, where the model is trained to estimate the pose for a specific object. However, these methods cannot generalize to other objects. Consequently, the category-level pose estimation~\cite{wang2019normalized} is introduced, where  the model is trained to estimate the pose for different instances within the same category. 
Nevertheless, they can hardly generalize to unseen objects of other categories, limiting their application potentials.

\begin{figure}[t]
    \centering
    \includegraphics[width=1\linewidth]{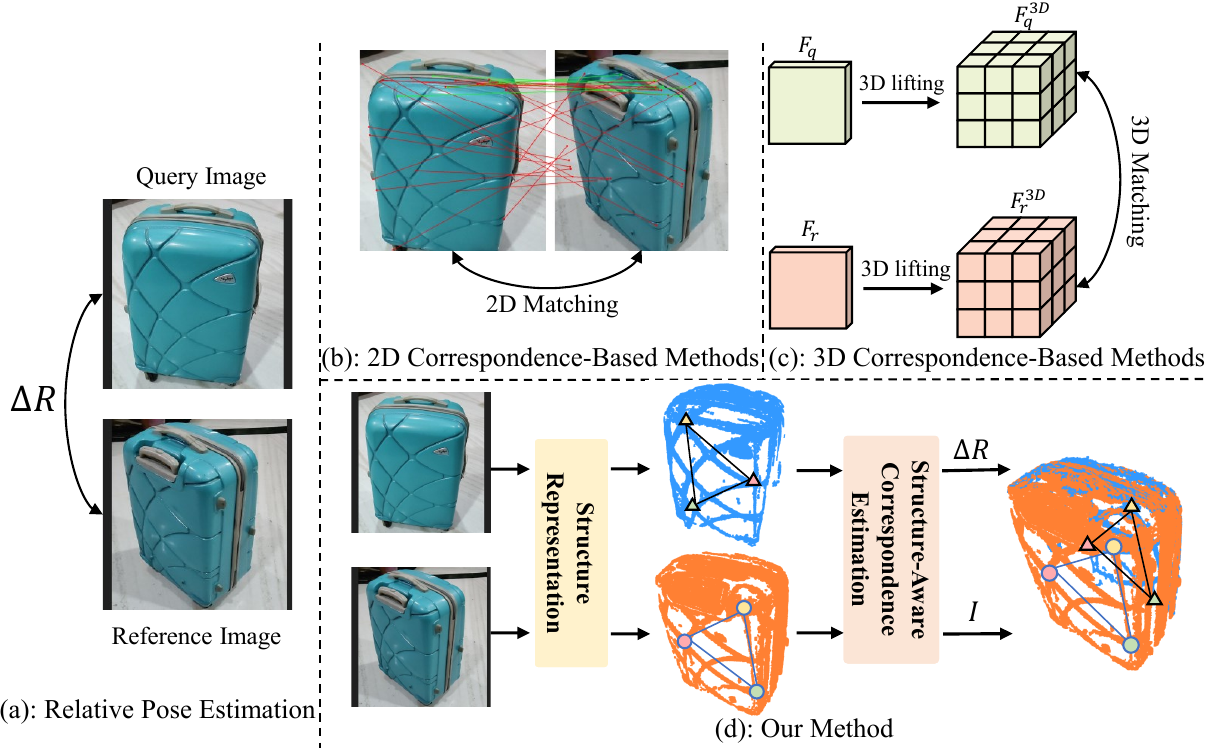}
    \caption{a) \textbf{Task depiction}. Predict the relative pose $\Delta R$ between the query and reference images. b) \textbf{2D correspondence-based methods} extract keypoints to conduct 2D-2D matching for pose estimation. c) \textbf{3D correspondence-based methods} lift 2D features into 3D voxel features and conduct 3D-3D matching for pose estimation. d) \textbf{Our method} bypasses the feature matching and directly regresses 3D correspondences for pose estimation.}
    \label{fig:1}
    \vspace{-10pt}
\end{figure}

To further improve the generalization ability, recent works~\cite{zhang2022relpose,lin2024relpose++,zhao20233d} have shifted towards relative pose estimation for unseen objects. It requires only a single reference image of a novel object to estimate the poses of new images, as shown in Figure \ref{fig:1} (a), providing a promising way to achieve object-agnostic  pose estimation. 
Existing relative pose estimation methods can be broadly categorized into three types, 2D correspondence-based methods~\cite{sun2021loftr,sarlin2020superglue}, hypothesis-and-verification-based methods~\cite{zhang2022relpose,lin2024relpose++,zhao20233d} and 3D correspondence-based methods~\cite{zhao20233d,zhao2024dvmnet}. 
2D correspondence-based methods~\cite{sun2021loftr,sarlin2020superglue} extract keypoints directly from two images and compute the relative pose based on keypoint matching. However, as shown in Figure \ref{fig:1} (b), these methods suffer from small overlapping regions caused by large pose variations, making it difficult to establish reliable correspondence. 
Hypothesis-and-verification-based methods~\cite{zhang2022relpose,lin2024relpose++,zhao20233d} sample a large number of pose hypotheses and then evaluate the score of each hypothesis through global feature matching or score regression. Nevertheless, these methods rely on a discrete sampling process and fail to adequately model the continuous pose space, which can only account for coarse pose estimation. Moreover, the verification of numerous pose hypotheses incurs significant computational costs. 
3D correspondence-based methods~\cite{zhao20233d,zhao2024dvmnet} lift 2D features into 3D voxel features and then conduct 3D-3D matching to estimate the pose, as shown in Figure \ref{fig:1} (c). Notably, these methods can establish correspondences even in invisible regions, showing promising potential. 
However, it's difficult to infer 3D features of invisible regions from 2D surface features without extra information, resulting in unreliable 3D matching for invisible regions. Moreover, dense 3D-3D matching process incurs high computational costs due to the cubic complexity.

Different from existing methods that rely on explicit  feature matching, we take inspiration from how humans assemble two object parts that have small or no overlapping regions. As shown in Figure \ref{fig:1} (a), the query and reference images represent the front and back of a suitcase, with only a small overlap at the top. By considering structural details like shape, the position of the handles, and the color pattern, we humans can intuitively infer how these two parts should be assembled to form a complete suitcase. This process of mentally assembling parts is actually akin to determining their relative pose. Inspired by this intuitive assembly process, we design a framework that leverages such object structural details to map keypoints from the query image into the reference 3D coordinate space, thus naturally establishing 3D-3D correspondences without explicit feature matching, as illustrated in Figure \ref{fig:1} (d).
Nevertheless, it is non-trivial because of the following challenges: (1) \textbf{How to represent the structure of each object part.} To reason how object parts in query image and reference image should be assembled, the model must understand the structure of these object parts. However, different objects or object parts captured in different views often exhibit significant variations in appearance and shape, making it challenge to design a method that can handle these situations well.
(2) \textbf{How to extract structure-aware features for correspondence estimation.} While humans can naturally assemble two object parts with complementary structures, it's non-trivial for the neural network. To accurately map points in the query image into the reference coordinate space,  it is crucial to encode the structure information of parts in query and reference image into the point features.

Based on the above discussion, we propose a Structure-Aware Correspondence Learning method for Relative Pose Estimation, which consists of a structure-aware keypoint extraction module and a structure-aware correspondence estimation module. Our key insight is to represent the structure of different parts through a set of keypoints and extract structure-aware keypoint features for correspondence estimation.  
\textbf{The structure-aware keypoint extraction module} is designed  to locate a set of sparse keypoints that can well represent the structure of different object parts. 
Specifically, to deal with the significant shape and appearance variations,  we use a set of learnable queries to interact with image features to produce image-specific keypoint detectors. We first compute similarities between keypoint detectors and image features to generate keypoint heatmaps, from which keypoint coordinates and features are derived. To guide the learning of the keypoint extraction module, we design an image reconstruction loss by constraining  keypoint features and coordinates to reconstruct the image. The intuition behind this loss is that if these keypoints represent the object structure well, the model can recover the original image from them.
\textbf{The structure-aware correspondence estimation module} is proposed to extract structure-aware keypoint features for correspondence estimation. To incorporate intra-image structure information, we use the self-attention mechanism that integrates relative keypoint positions to aggregate features from other keypoints in the same image. To incorporate inter-image structure information, we apply the cross-attention mechanism to aggregate keypoint features from the other image. Consequently, the extracted structure-aware keypoint features enables the network to perceive how the object parts in two images should be assembled, which can facilitate the correspondence estimation. Given these structured-aware keypoint features, we lift 2D keypoints to 3D space within the query coordinate system  and regress their corresponding 3D coordinates within the reference coordinate system to establish 3D-3D correspondences. Finally, we estimate the relative pose with 3D-3D correspondence by employing a weighted Singular Value Decomposition, ensuring end-to-end optimization.

In summary, our contributions are as follows:
\begin{itemize}
    \item We propose a novel structure-aware correspondence learning method for relative pose estimation, which can establish robust 3D-3D correspondence  without explicit feature matching.
    \item We propose two key designs, a structure-aware keypoint extraction module that can well represent the structure of different object parts, and a structure-aware correspondence estimation module that can help the keypoints to aggregate structure-aware features for robust correspondence estimation.
    \item Experimental results on three challenging datasets, CO3D~\cite{reizenstein2021common}, Objaverse~\cite{deitke2023objaverse} and LineMOD~\cite{hinterstoisser2013model}, demonstrate the state-of-the-art performance of our method.
\end{itemize}

\vspace{-5pt}
\section{Related Work}

\label{sec:related}
\vspace{-5pt}
\subsection{Instance-Level Object Pose Estimation}
\vspace{-5pt}
Instance-level pose estimation methods~\cite{xiang2017posecnn,wang2019densefusion,he2020pvn3d,he2021ffb6d} predict the 6D pose of specific known objects by leveraging CAD models. Recent approaches include correspondence, template, voting, and direct regression methods. Correspondence-based methods~\cite{rad2017bb8,li2019cdpn,zakharov2019dpod} establish matches between inputs and CAD models, then estimate pose via PnP~\cite{fischler1981random} or similar algorithms. Template-based methods~\cite{sundermeyer2018implicit,li2022dcl} match inputs to pre-defined templates, treating pose estimation as a matching problem. Voting-based methods~\cite{liu2021kdfnet,tian2020robust} aggregate votes from pixels or points, either by keypoints or direct pose prediction. Regression-based methods~\cite{gao20206d,lin20216d} directly predict 6D poses from images or depth data, simplifying the pipeline. Although effective for known objects, these methods struggle to generalize to unseen objects.

\subsection{Category-level Object Pose Estimation}

To extend pose estimation beyond specific instances, researchers have developed category-level methods~\cite{wang2019normalized,tian2020shape,lin2022category,lin2023vi,lin2024instance,iclr2025spherepose, cvpr2025spotpose} for estimating object poses within predefined categories without CAD models. These methods fall into two categories: shape prior-based and shape prior-free. Shape prior-based approaches~\cite{wang2019normalized,tian2020shape,lin2022category} utilize CAD-derived priors for alignment or direct regression of poses. In contrast, shape prior-free methods~\cite{lin2023vi,lin2024instance,chen2024secondpose,iclr2025spherepose, cvpr2025spotpose} remove reliance on priors, learning features directly from input data to estimate poses. Although category-level methods improve generalization over instance-specific methods, they still struggle to generalize to unseen categories.

\subsection{Relative Object Pose Estimation}

To enhance the generalization of pose estimation, recent work~\cite{zhang2022relpose,zhao20233d,lin2024relpose++,zhao2024dvmnet} has focused on relative pose estimation for unseen objects. Unlike instance-specific methods, these approaches require only a single reference image of a novel object to estimate the relative pose of image, making them highly suitable for applications where data acquisition is costly. Current methods can be broadly categorized into three types: 2D correspondence-based methods~\cite{sun2021loftr,sarlin2020superglue,lowe2004distinctive}, hypothesis-and-verification-based methods~\cite{zhang2022relpose,lin2024relpose++,zhao20233d,nguyen2024nope}, and 3D correspondence-based methods~\cite{zhao20233d,zhao2024dvmnet}. 2D correspondence-based methods~\cite{sun2021loftr,sarlin2020superglue,lowe2004distinctive} establish correspondences between keypoints from both images and use these to compute relative pose. Learned feature-based approaches like SuperGlue~\cite{sarlin2020superglue} and LoFTR~\cite{sun2021loftr} have achieved robustness in feature matching under moderate viewpoint changes and lighting variations. However, with only a single reference image and substantial viewpoint differences, these methods struggle due to their sensitivity, significantly affecting accuracy~\cite{zhang2022relpose, zhao20233d}. Hypothesis-and-verification-based methods~\cite{zhang2022relpose,lin2024relpose++,zhao20233d,nguyen2024nope} mitigate these challenges by generating multiple pose hypotheses over the rotation space and evaluating them through similarity networks, as in RelPose~\cite{zhang2022relpose} and RelPose++~\cite{lin2024relpose++}. While effective in handling larger viewpoint differences, these methods require extensive sampling and verification, resulting in high computational costs that restrict their suitability for real-time applications. As an alternative, 3D correspondence-based methods~\cite{zhao2024dvmnet,zhao20233d} lift 2D features into 3D voxel features and then conduct 3D-3D matching to estimate the pose. For example, DVMNet~\cite{zhao2024dvmnet} uses lifted 3D voxel features to facilitate matching even in invisible regions. However, it’s difficult to infer 3D features solely from 2D surface image features, which leads to unreliable 3D matching. 
Moreover, the dense 3D-3D matching incurs high computational costs due to the cubic complexity.

\section{Method}
\label{sec:method}

\begin{figure*}[t]
    \centering
    \includegraphics[width=1\linewidth]{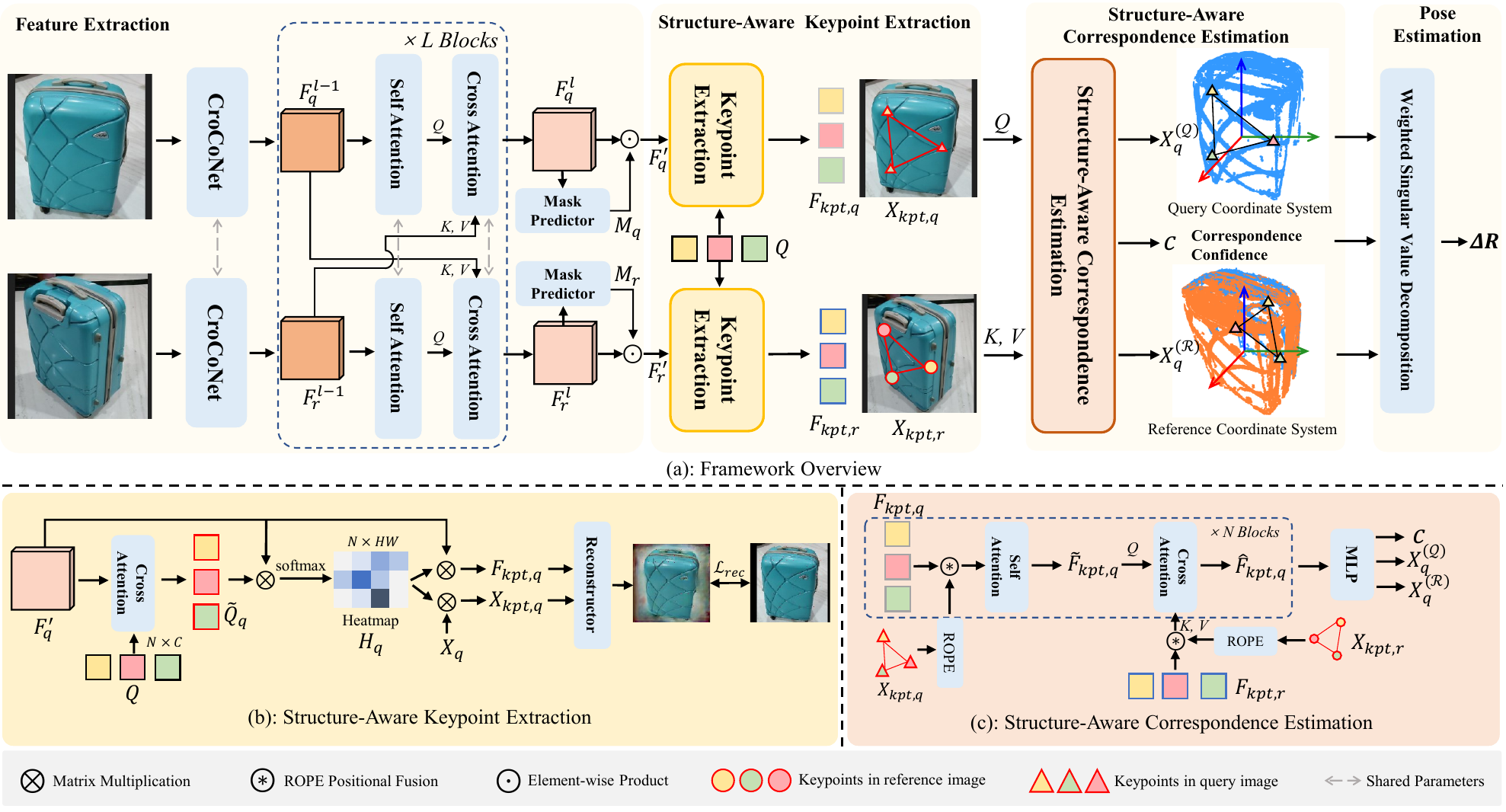}
    \caption{a) Overview of the proposed method. b) Illustration of the Structure-Aware Keypoint Extraction module. We initialize a set of learnable queries that interact with image features to extract keypoints representing the object's structure. And we further employ a reconstructor and $\mathcal{L}_{\text{rec}}$ for supervision of keypoints extraction. c) Illustration of the Structure-Aware Correspondence Estimation module. We employ ROPE and an attention mechanism to extract structure-aware features for 3D correspondence estimation. }
    \label{fig:pipeline}
\end{figure*}

\subsection{Overview}

We tackle the problem of estimating the relative pose between a query image and a reference image belonging to previously unseen object categories. Specifically, we denote the set of object categories available during training as $\omega_{\text{train}}$ and the set of categories used during testing as $\omega_{\text{test}}$, where $\omega_{\text{train}} \cap \omega_{\text{test}} = \emptyset$. This setup introduces several key challenges, including significant viewpoint variations between the images, minimal overlapping regions, and the generalization to novel object categories. Since the translation component of 6D pose can be reliably estimated through existing 2D detection techniques~\cite{he2017mask,kirillov2023segment}, we focus on addressing the more challenging 3D rotation estimation, which is the same with the mainstream methods~\cite{zhang2022relpose,zhao20233d,zhao2024dvmnet}.

The framework of our method is shown in Figure \ref{fig:pipeline} (a).
Given the query image $I_q$ and the reference image $I_r$, we first employ a shared feature extractor to obtain feature maps $\mathbf{F}_{q}$ and $\mathbf{F}_{r}$. These feature maps are subsequently passed through symmetric attention blocks~\cite{vaswani2017attention} to enhance the image features mutually. Afterward, we use the proposed structure-aware keypoint extraction module to extract keypoints independently from each image. Then, by leveraging our proposed structure-aware correspondence estimation module, we facilitate both intra-image and inter-image feature interactions for the keypoints extracted from the query image, enabling structure-aware feature aggregation. With these updated keypoint features, we lift 2D keypoints to 3D space and regress their corresponding 3D coordinates within the reference coordinate system. Finally, based on the established correspondences, the relative rotation $\Delta \mathbf{R}$ is obtained via a weighted Singular Value Decomposition (wSVD) algorithm~\cite{besl1992method}.

\subsection{Feature Extraction}

We first utilize a pre-trained backbone~\cite{weinzaepfel2023croco} to extract image features from both the query and reference images, yielding feature maps $\mathbf{F}_{q}, \mathbf{F}_{r} \in \mathbb{R}^{H \times W \times C}$. These feature maps are then processed through a two-step attention mechanism with Multi-Head Self-Attention (MHSA) and Multi-Head Cross-Attention (MHCA) layers to update features:
\begin{equation}
\begin{aligned}
\tilde{\mathbf{F}}_{q}^{(l-1)} &= \text{MHSA}(\mathbf{F}_{q}^{(l-1)}) + \mathbf{F}_{q}^{(l-1)}, \\
\mathbf{F}_{q}^{(l)} &= \text{MHCA}(\tilde{\mathbf{F}}_{q}^{(l-1)}, \mathbf{F}_{r}^{(l-1)}) + \tilde{\mathbf{F}}_{q}^{(l-1)}.
\end{aligned}
\label{eq:cross_attention}
\end{equation}
The same operations are performed for $\mathbf{F}_{r}$. Unlike previous works~\cite{zhao20233d,zhao2024dvmnet}, the parameters in the MHSA and MHCA modules are shared across the query ($q$) and reference ($r$) images, ensuring consistency between the learned features. After $L$ layers of such interactions, we obtain the final feature maps $\mathbf{F}_{q}^{(L)}$ and $\mathbf{F}_{r}^{(L)}$. 

To suppress the influence of background features, we apply a lightweight mask predictor to obtain an object mask to refine the feature maps. Formally, the mask $\mathbf{M}_{q}$ and $\mathbf{M}_{r}$ are derived from $\mathbf{F}_{q}^{(L)}$ and $\mathbf{F}_{r}^{(L)}$, respectively:
\begin{equation}
\begin{aligned}
\mathbf{M}_{q} &= g(\mathbf{F}_{q}^{(L)}), \\
\mathbf{M}_{r} &= g(\mathbf{F}_{r}^{(L)}).
\end{aligned}
\label{eq:mask}
\end{equation}

We use binary cross-entropy (BCE) loss and ground truth masks to compute the mask prediction loss as follows:
\begin{equation}
\mathcal{L}_{\text{mask}} = \text{BCE} (\mathbf{M}_{q}, \mathbf{M}_{q}^{\text{gt}}) + \text{BCE} (\mathbf{M}_{r}, \mathbf{M}_{r}^{\text{gt}})
\end{equation}
The final feature maps are obtained by element-wise multiplication with these object masks, effectively retaining only object-relevant features and reducing interference from the background:
\begin{equation}
\begin{aligned}
\mathbf{F}_{q}^{\prime} &= \mathbf{F}_{q}^{(L)} \odot \mathbf{M}_{q}, \\
\mathbf{F}_{r}^{\prime} &= \mathbf{F}_{q}^{(L)} \odot \mathbf{M}_{r}.
\end{aligned}
\label{eq:mask_refinement}
\end{equation}

\subsection{Structure-Aware Keypoint Extraction}

As introduced in Section~\ref{sec:intro}, a key challenge is how to effectively represent the structure of object parts, as they often exhibit significant variations in appearance and shape.
To address these issues, we propose the structure-aware keypoint extraction module to adaptively select keypoints with structural significance, as illustrated in Figure \ref{fig:pipeline} (b). 

In the following, we use the query image as an example to illustrate the keypoint detection process. 
Specifically, we initialize a set of learnable queries, denoted as $\mathbf{Q} \in \mathbb{R}^{N_{\text{kpt}} \times C}$, where $N_{\text{kpt}}$ is the number of keypoints, and $C$ is the feature dimension. 
To convert the this queries into image-specific keypoint detectors $\tilde{\mathbf{Q}}_q$,  we again use the attention mechanism to update these queries with the image features, so as to adapt to the content of different images.
\begin{equation}
\tilde{\mathbf{Q}}_q = \text{MHCA}(\mathbf{Q}, \mathbf{F}_{q}^{\prime}) + \mathbf{Q}.
\end{equation}
Next, we compute the similarity between these image-specific keypoint detectors and the image features, generating keypoints heatmap $\mathbf{H}_{q} \in \mathbb{R}^{N_{\text{kpt}} \times H \times W}$:
\begin{equation}
\mathbf{H}_{q} = \text{softmax}\left(\tilde{\mathbf{Q}}_q \cdot {\mathbf{F}_{q}^{\prime}}^{\top}\right).
\end{equation}

After that, we derive the spatial coordinates and the corresponding features for all keypoints by performing a weighted averaging based on the heatmap $\mathbf{H}_{q}$:
\begin{align}
\mathbf{X}_{kpt,q} &= \sum_{h=1}^H \sum_{w=1}^W \mathbf{H}_{q}(h, w) \cdot (h, w), \\
\mathbf{F}_{kpt,q} &= \sum_{h=1}^H \sum_{w=1}^W \mathbf{H}_{q}(h, w) \cdot \mathbf{F}_{q}^{\prime}(h, w),
\end{align}
where $\mathbf{X}_{kpt,q} \in \mathbb{R}^{N_{\text{kpt}} \times 2 }$ denotes the spatial coordinates of all keypoints, and $\mathbf{F}_{kpt,q} \in \mathbb{R}^{N_{\text{kpt}} \times C }$ denotes their corresponding features.
However, without explicit constraints, the extracted keypoints often cluster within limited regions, reducing their effectiveness in capturing comprehensive structural information of the object part. 
To address this, we introduce an image reconstruction loss that drives keypoints to cover semantically rich regions of the object by reconstructing its foreground solely from the keypoint features and coordinates.
\begin{equation}
\hat{\mathbf{I}}_{q} = f(\mathbf{X}_{\text{kpt},q}, \mathbf{F}_{\text{kpt},q}),
\end{equation}
where $f(\cdot, \cdot)$ is a lightweight decoder, and $\hat{\mathbf{I}}_{q}$ denotes the reconstructed query image.
The reconstruction loss consists of an $L_2$ loss for pixel-wise similarity and a VGG-based perceptual loss:
\begin{align}
\mathcal{L}_{\text{rec}, q} &= \lambda_1 \lVert \hat{\mathbf{I}}_{q} - \mathbf{I}_{q} \rVert_2^2 + \lambda_2 \sum_{l} \lVert \phi_l(\hat{\mathbf{I}}_{q}) - \phi_l(\mathbf{I}_{q}) \rVert_2^2,
\end{align}
where $\phi_l(\cdot)$ denotes the feature map from the $l$-th layer of the VGG network~\cite{simonyan2014very}, and $\lambda_1, \lambda_2$ are weighting factors. The $L_2$ loss ensures pixel accuracy, while the perceptual loss maintains semantic consistency. By reconstructing the image, we can optimize keypoint distribution end-to-end, ensuring these keypoints cover semantically rich regions of the object's surface and enhance structural representation.

By applying the above process to both the query and reference images, we obtain structurally significant keypoints that effectively represent the object's structure.

\subsection{Structure-Aware Correspondence Estimation}

Given the 2D keypoint coordinates $\mathbf{X}_{kpt,q}$, $\mathbf{X}_{kpt,r}$ and associated features $\mathbf{F}_{kpt,q}$, $\mathbf{F}_{kpt,r}$ from both the query and reference images, we extract structure-aware features to lift 2D keypoints to 3D space within the query coordinate system and regress their corresponding 3D coordinates within the reference coordinate system, establishing a set of 3D correspondences for relative pose estimation.

Specifically, for keypoint features $\mathbf{F}_{\text{kpt},q} \in \mathbb{R}^{N_{\text{kpt}} \times C}$ extracted from the query image, we refine these features using self-attention with rotational positional encoding (ROPE)~\cite{su2024roformer}, enabling the keypoint features to perceive the intra-image structure. Here, we denote the ROPE encoding as $R(\cdot)$, and $\circledast$ indicates the ROPE positional fusion operation as used in the original method~\cite{su2024roformer}.

\begin{equation}
\tilde{\mathbf{F}}_{\text{kpt},q} = \text{MHSA}(\mathbf{F}_{\text{kpt},q}\circledast {R}(\mathbf{X}_{\text{kpt},q})).
\end{equation}
We then apply cross-attention to aggregate structure information from reference image, where the refined keypoint features from the query image act as queries, and the reference keypoint features are used as keys and values:
\begin{equation}
\hat{\mathbf{F}}_{\text{kpt},q} = \text{MHCA}\left( \tilde{\mathbf{F}}_{\text{kpt},q} \circledast {R}(\mathbf{X}_{\text{kpt},q}), \, \mathbf{F}_{\text{kpt},r} \circledast {R}(\mathbf{X}_{\text{kpt},q}) \right).
\end{equation}

These attention mechanisms help capture intra- and inter-image relationships, enabling the keypoint features for robust 3D correspondence estimation.
With the updated keypoint features, we lift the 2D keypoints into 3D space within the query coordinate system by regressing a pseudo-depth value $d_{i, q}$ for each keypoint:
\begin{equation}
d_{i, q} = \text{MLP}_{\text{depth}}(\hat{\mathbf{f}}_{i, q}),
\end{equation}
where $d_{i, q}$ denotes the pseudo-depth of the $i$-th keypoint in the query image, and $\hat{\mathbf{f}}_{i, q}$ denotes the $i$-th keypoint feature from the updated keypoint feature set $\hat{\mathbf{F}}_{\text{kpt},q}$. By concatenating this depth value with the corresponding 2D coordinates, we obtain the 3D coordinates for each keypoint in the query coordinate system:
\begin{equation}
\mathbf{x}_{i, q}^{(\mathcal{Q})} = \left[\mathbf{x}_{i, q}, d_{i, q}\right] \in \mathbb{R}^3.
\end{equation}
The superscript $(\mathcal{Q})$ indicates 3D coordinates in the query system and subscript $q$ indicates the keypoint extracted from the query image. The notation $(\mathcal{R})$ and $r$ represent the reference correspondingly.
Similarly, we estimate the corresponding 3D coordinates for the keypoints in the reference coordinate system using another MLP, which takes the updated keypoint feature $\hat{\mathbf{f}}_{i, q}$ and corresponding 3D coordinate in the query system as input:
\begin{equation} 
\mathbf{x}_{i, q}^{(\mathcal{R})}, c_{i} = \text{MLP}_{\text{ref}}([\hat{\mathbf{f}}_{i, q}, PE(\mathbf{x}_{i, q}^{(\mathcal{Q})})]),
\end{equation}
where $\mathbf{x}_{i, q}^{(\mathcal{R})}$ denotes the estimated 3D coordinates of the $i$-th keypoint in the reference coordinate system, and $c_{i} \in [0, 1]$ denotes the confidence score of this keypoint.
Based on the 3D coordinates of the same keypoints within both the query and reference systems, we can naturally establish 3D-3D correspondences, which helps determine the relative pose.
%
To ensure the accuracy of these correspondences, we propose a loss function that supervises the predicted 3D coordinates.
Specifically, the ground truth 3D coordinates in the reference system are computed using the ground truth relative rotation matrix $\Delta \textbf{R}_{\text{gt}}$:
\begin{equation}
\mathbf{x}_{i, q,\text{gt}}^{(\mathcal{R})} = \Delta \textbf{R}_{\text{gt}} \cdot \mathbf{x}_{i, q}^{(\mathcal{Q})}.
\end{equation}
%
To align the predicted 3D coordinates $\mathbf{x}_{i, q}^{(\mathcal{R})}$ with these ground truth values, we define the 3D keypoint loss $\mathcal{L}_{\text{pts}}$ as follows:
\begin{equation}
\mathcal{L}_{\text{pts}} = \frac{1}{N_{\text{kpt}}} \sum_{i=1}^{N_{\text{kpt}}} \left( c_{i} \cdot e_{i, r} - \alpha \log(c_{i}) \right),
\end{equation}
where $c_{i}$ denotes the confidence score of each keypoint, and $\alpha$ is a hyperparameter controlling the influence of the confidence score. The 3D keypoint error $e_{i, r}$ measures the discrepancy between the estimated and ground truth coordinates:
\begin{equation}
e_{i, r} =  \frac{1}{2} \left( \lVert \mathbf{x}_{i, q}^{(\mathcal{R})} - \text{sg}(\mathbf{x}_{i, q,\text{gt}}^{(\mathcal{R})}) \rVert_2^2 + \lVert \text{sg}(\mathbf{x}_{i, q}^{(\mathcal{R})}) - \mathbf{x}_{i, q,\text{gt}}^{(\mathcal{R})} \rVert_2^2 \right),
\end{equation}
where $\text{sg}(\cdot)$ denotes the stop-gradient operation. This symmetric loss penalizes deviations in both coordinate systems, ensuring the 3D coordinates $\mathbf{x}_{i, q}^{(\mathcal{Q})}$ and $\mathbf{x}_{i, q}^{(\mathcal{R})}$ are accurately estimated, ultimately leading to reliable 3D correspondences.

\subsection{Pose Estimation via 3D Correspondences}

After obtaining the 3D keypoint correspondences, following previous work~\cite{zhao2024dvmnet}, we employ a weighted Singular Value Decomposition (wSVD) approach to solve the relative rotation $\Delta \textbf{R}$:
\begin{equation}
\Delta \textbf{R} = \arg \min_{\mathbf{R}} \sum_{i=1}^{N_q} c_{i} \lVert \mathbf{x}_{i, q}^{(\mathcal{R})} - \mathbf{R} \mathbf{x}_{i, q}^{(\mathcal{Q})} \rVert_2^2,
\end{equation}
where $\mathbf{x}_{i, q}^{(\mathcal{R})}$ and $\mathbf{x}_{i, q}^{(\mathcal{Q})}$ are the 3D coordinates of the $i$-th keypoint in the query and reference coordinate systems, and $c_{i}$ denotes corresponding confidence score estimated earlier. The optimization seeks the rotation matrix $\Delta \textbf{R}$ that best aligns these two sets of 3D coordinates.

To solve the minimization problem for estimating the optimal rotation matrix $\Delta \textbf{R}$, we follow the approach of utilizing SVD method to derive the optimal alignment between two sets of 3D keypoints, as used in prior works on point cloud registration~\cite{arun1987least}. Specifically, we first compute the covariance matrix $\mathbf{H}$ using the query and reference 3D keypoints along with their confidence scores:
\begin{equation}
\mathbf{H} = \sum_{i=1}^{N_q} c_{i} \mathbf{x}_{i, q}^{(\mathcal{Q})} (\mathbf{x}_{i, q}^{(\mathcal{R})})^\top.
\end{equation}
We then perform Singular Value Decomposition (SVD) on the covariance matrix $\mathbf{H}$:
\begin{equation}
\mathbf{H} = \mathbf{U} \mathbf{\Sigma} \mathbf{V}^\top.
\end{equation}
The optimal relative rotation $\Delta \textbf{R}$ is then obtained as:
\begin{equation}
\Delta \mathbf{R} = \mathbf{V} \mathbf{U}^\top.
\end{equation}
To align the estimated relative rotation $\Delta \mathbf{R}$ with the ground truth $\Delta \mathbf{R}_{\text{gt}}$, we employ the $L_1$ loss: 
\begin{equation}
\mathcal{L}_{\text{rot}} = \lVert q(\Delta \mathbf{R}) - q(\Delta \mathbf{R}_{\text{gt}}) \rVert_1,
\end{equation}
where $q(\Delta \mathbf{R})$ and $q(\Delta \mathbf{R}_{\text{gt}})$ denote the 6D representation for $\Delta \mathbf{R}$ and $\Delta \mathbf{R}_{\text{gt}}$ introduced in ~\cite{zhou2019continuity}.

\subsection{Training and Inference Details}

Our framework optimizes a combined loss function during training, which includes the 3D keypoint loss $\mathcal{L}_{\text{pts}}$, the reconstruction loss $\mathcal{L}_{\text{rec}}$, the rotation loss $\mathcal{L}_{\text{rot}}$, and the mask loss $\mathcal{L}_{\text{mask}}$. The total loss can be expressed as:
\begin{equation}
\mathcal{L}_{\text{total}} = \lambda_1 \mathcal{L}_{\text{pts}} + \lambda_2 \mathcal{L}_{\text{rec}} + \lambda_3 \mathcal{L}_{\text{rot}} + \lambda_4 \mathcal{L}_{\text{mask}},
\end{equation}
where $\lambda_1$, $\lambda_2$, $\lambda_3$, and $\lambda_4$ are hyperparameters controlling the influence of each term.
Additionally, during training, both query and reference images are used symmetrically. Specifically, the keypoints extracted from the reference image are also used to obtain a set of 3D correspondences, effectively providing additional training data that enhances efficiency and robustness. During inference, the model extracts keypoints from both query and reference images, but only regresses the 3D coordinates from the query image for the relative pose estimation.


\section{Experiment}
\label{sec:experiment}

\begin{table*}[htbp]
    \centering
    \setlength{\tabcolsep}{2pt} 
    \small 
    \renewcommand{\arraystretch}{1.0} 
    \caption{Performance comparison with state-of-the-art methods on CO3D, Objaverse, and LineMOD datasets. Here we denote 2D correspondence-based methods as 2D, hypothesis-and-verification-based methods as H\&V, 3D correspondence-based methods as 3D.}
    \vspace{-8pt}
    \begin{tabular}{l|c|ccc|ccc|ccc}
        \toprule
        \multirow{2}{*}{\textbf{Method}} & \multirow{2}{*}{\textbf{Type}} 
        & \multicolumn{3}{c|}{\textbf{CO3D}} 
        & \multicolumn{3}{c|}{\textbf{Objaverse}} 
        & \multicolumn{3}{c}{\textbf{LineMOD}} \\
        & & mAE $\downarrow$ & Acc@30$^\circ$ $\uparrow$ & Acc@15$^\circ$ $\uparrow$ 
        & mAE $\downarrow$ & Acc@30$^\circ$ $\uparrow$ & Acc@15$^\circ$ $\uparrow$ 
        & mAE $\downarrow$ & Acc@30$^\circ$ $\uparrow$ & Acc@15$^\circ$ $\uparrow$ \\
        \midrule
        SuperGlue~\cite{sarlin2020superglue} & 2D & 67.2 & 45.2 & 37.7 & 102.4 & 15.1 & 12.1 & 64.8 & 26.2 & 14.3 \\
        LoFTR~\cite{sun2021loftr} & 2D & 77.5 & 37.9 & 33.1 & 134.1 & 9.6 & 7.7 & 84.5 & 24.2 & 13.5 \\
        ZSP~\cite{goodwin2022zero} & 2D & 87.5 & 25.7 & 14.6 & 107.2 & 4.2 & 1.5 & 78.6 & 10.7 & 2.7 \\
        RelPose~\cite{zhang2022relpose} & H\&V & 50.0 & 64.2 & 48.6 & 80.4 & 20.8 & 6.7 & 58.3 & 26.1 & 7.0 \\
        RelPose++~\cite{lin2024relpose++} & H\&V & 38.5 & 77.0 & 69.8 & 33.5 & 72.3 & 42.9 & 46.6 & 42.5 & 15.1 \\
        3DAHV~\cite{zhao20233d} & H\&V 3D & 28.5 & 83.5 & 71.0 & 28.1 & 78.6 & 58.4 & 41.7 & 61.5 & 29.9 \\
        DVMNet~\cite{zhao2024dvmnet} & 3D & 19.9 & 85.9 & 62.3 & 20.2 & 81.5 & 57.2 & 36.8 & 55.1 & 23.8 \\
        \midrule
        \textbf{Ours} & 3D & \textbf{14.2} & \textbf{93.6} & \textbf{80.2} & \textbf{15.3}& \textbf{90.3} & \textbf{74.0}  &\textbf{27.2} & \textbf{76.2} & \textbf{41.8} \\
        \bottomrule
    \end{tabular}
    \label{tab:comparison}
    \vspace{-8pt}
\end{table*}

\textbf{Datasets.} Following previous works~\cite{zhao20233d, zhao2024dvmnet}, we evaluate our method on CO3D~\cite{reizenstein2021common}, Objaverse~\cite{deitke2023objaverse} and LineMOD~\cite{hinterstoisser2013model}, which are widely-used datasets for relative pose estimation. These datasets include diverse synthetic and real data across various object categories.
The CO3D dataset contains 18,619 video sequences spanning 51 categories. Following~\cite{reizenstein2021common}, we train on 41 categories and test on 10 unseen categories to evaluate generalization. 
The Objaverse dataset consists of synthetic images rendered from 3D models across diverse viewpoints. We select 128 objects for testing and reserve the remaining for training.
For LineMOD, we use calibrated real images of 13 household objects. The test set includes 5 objects, which are excluded from training to ensure complete separation.

\textbf{Implementation Details.} The Adam optimizer~\cite{kingma2014adam} is employed with an initial learning rate of $2 \times 10^{-4}$, which decays by a factor of 0.1 every 200 epochs. The model is trained for 400 epochs with a batch size of 80. All experiments are conducted on 4 NVIDIA RTX3090 GPUs, taking roughly 36 hours. Following previous works~\cite{zhang2022relpose,lin2024relpose++,zhao20233d,zhao2024dvmnet}, we crop the object from the image by utilizing the ground truth bounding box. Further details can be found in the supplementary material.

\textbf{Evaluation Metrics.} Following previous works~\cite{zhao20233d, zhao2024dvmnet}, we evaluate our model using two metrics: mean angular error (mAE) and accuracy under predefined thresholds. The angular error $\theta$ between the predicted rotation $\Delta R$ and the ground truth $\Delta R_{\text{gt}}$ is calculated as:

\begin{equation}
\theta = \arccos \left( \frac{\text{Tr}(\Delta R_{\text{gt}}^\top \Delta R) - 1}{2} \right),
\end{equation}
where $\Delta R_{\text{gt}}$ and $\Delta R$ are the ground truth and predicted rotation matrices, respectively.
We also report accuracy as the percentage of test samples with an angular error below thresholds of $30^\circ$ and $15^\circ$. 

\subsection{Comparison with State-of-the-Art Methods}

Table \ref{tab:comparison} presents a comprehensive comparison between our method and state-of-the-art methods on the CO3D, Objaverse, and LineMOD datasets. Our method consistently outperforms prior methods across all datasets and metrics.

Our approach significantly outperforms traditional 2D feature-based methods, such as SuperGlue~\cite{sarlin2020superglue} and LoFTR~\cite{sun2021loftr}, primarily due to their inability to reliably match keypoints under large pose differences and minimal overlap areas. Furthermore, our method achieves superior results compared to hypothesis-and-verification-based methods like RelPose~\cite{zhang2022relpose} and RelPose++~\cite{lin2024relpose++}. These methods rely on global features while ignoring local structural cues, and their use of discrete sampling limits their ability to model the continuous pose space accurately.

Compared to the 3D correspondence-based methods, such as DVMNet~\cite{zhao2024dvmnet} and 3DAHV~\cite{zhao20233d}, our method demonstrates substantial improvements. On the CO3D dataset, our approach reduces mean angular error (mAE) by nearly $6^\circ$, and improves $\text{Acc @ } 30^\circ$ and $\text{Acc @ } 15^\circ$ by approximately 8$\%$. 
These gains are primarily because it is difficult to infer reliable 3D features without extra information, often leading to incorrect matches and unreliable 3D correspondences . In contrast, our approach avoids the matching process and directly regresses accurate 3D correspondences using structural information.

Our method also achieves state-of-the-art results on the Objaverse and LineMOD datasets, which include a variety of synthetic and real-world object categories. The observed improvements in mAE and accuracy on these datasets demonstrate that the proposed method can generalize well to diverse conditions, providing a promising  solution for real-world applications.

\begin{table}[t]
    \centering
    \caption{Ablation study on the effectiveness of the Structure-Aware Keypoint Extraction module.}
    \vspace{-8pt}
    \renewcommand{\arraystretch}{1}
    \setlength{\tabcolsep}{6pt}
    \small 
    \setlength{\tabcolsep}{4pt} 
    \begin{tabular}{c|c|c|c|c}
        \toprule
        \text{Setting} & mAE $\downarrow$ & Acc@30$^\circ$ $\uparrow$ & Acc@15$^\circ$ $\uparrow$ &MACs(G)  \\
        \midrule
        \text{Dense} & 15.52   & 92.65  & 78.20   & 55.26 \\
        \text{Random} & 20.15 & 88.34  & 68.99  & 49.59\\
        \text{Keypoint} & \textbf{14.2} & \textbf{93.6} & \textbf{80.2} & 50.05 \\
        \bottomrule
    \end{tabular}
    \label{tab:ablation1}
    \vspace{-8pt}
\end{table}

\vspace{-2.5pt}
\subsection{Ablation Studies}

In this section, we conduct ablation studies to demonstrate the effectiveness of each design on the CO3D dataset.

\textbf{Effects of the Structure-Aware Keypoint Extraction.} To evaluate the effectiveness of our Structure-Aware Keypoint Extraction module, we conducted ablation experiments by replacing extracted keypoints with dense pixel features or randomly sampled points. As shown in Table \ref{tab:ablation1}, our module consistently outperforms both alternatives across all metrics. The results show that our keypoint extraction module effectively captures object structure, even with significant shape and appearance variations. In contrast, dense pixel features introduce irrelevant background noise or insignificant features and incur high computational costs, while random sampling lacks consistency in representing object structure. In summary, our method yields superior performance with fewer Multiply-Accumulate Operations (MACs), demonstrating efficiency and effectiveness.

\begin{table}[htbp]
    \centering
    \caption{Ablation study on the effectiveness of the Structure-Aware Correspondence Estimation module.}
    \vspace{-7.5pt}
    \renewcommand{\arraystretch}{1}
    \setlength{\tabcolsep}{6pt}
    \begin{tabular}{c|c|c|c}
        \toprule
        \text{Setting} & mAE $\downarrow$ & Acc@30$^\circ$ $\uparrow$ & Acc@15$^\circ$ $\uparrow$ \\
        \midrule
        \text{w/o Self} & 17.79  & 90.22 & 72.48  \\
        \text{w/o Cross} & 18.53  & 89.38 & 70.99   \\
        \text{Dense Reg.} & 19.37 & 88.51 & 65.17 \\
        \text{Global Reg.} & 21.97 & 86.76  & 68.16\\
        \text{Ours}& \textbf{14.2} & \textbf{93.6} & \textbf{80.2} \\
        \bottomrule
    \end{tabular}
    \label{tab:ablation2}
    \vspace{-7.5pt}
\end{table}

\begin{table}[htbp]
    \centering
    \caption{Ablation studies on the mask loss and confidence score.}
    \vspace{-7.5pt}
    \renewcommand{\arraystretch}{1}
    \setlength{\tabcolsep}{6pt}
    \begin{tabular}{c|c|c|c}
        \toprule
        \text{Setting} & mAE $\downarrow$ & Acc@30$^\circ$ $\uparrow$ & Acc@15$^\circ$ $\uparrow$ \\
        \midrule
        \text{w/o $\mathcal{L}_{\text{mask}}$} & 16.37 & 91.68 & 76.82   \\
        \text{w/o confidence} & 15.58 & 92.02   & 77.41  \\
        \text{Ours} & \textbf{14.2} & \textbf{93.6} & \textbf{80.2} \\
        \bottomrule
    \end{tabular}
    \label{tab:ablation3}
    \vspace{-7.5pt}
\end{table}

\begin{table}[htbp]
    \centering
    \caption{Ablation study on the number of keypoints.}
    \vspace{-7.5pt}
    \renewcommand{\arraystretch}{1}
    \setlength{\tabcolsep}{6pt}
    \small 
    \begin{tabular}{c|c|c|c|c}
        \toprule
        $N_{\text{kpt}}$ & mAE $\downarrow$ & Acc@30$^\circ$ $\uparrow$ & Acc@15$^\circ$ $\uparrow$ & MACs(G) \\
        \midrule
        16 & 18.03 & 90.38  & 72.99 & 48.52 \\
        32 & 15.95      & 92.30  &  76.54&  49.28\\
        \textbf{48} & \textbf{14.2} & 93.6 & \textbf{80.2}& 50.05\\
        64 & 14.35   & \textbf{93.85} & 79.67& 50.82 \\
        \bottomrule
    \end{tabular}
    \label{tab:ablation4}
    \vspace{-10pt}
\end{table}

\textbf{Effects of the Structure-Aware Correspondence Estimation.} To evaluate the effectiveness of the proposed Structure-Aware Correspondence Estimation module, we conducted ablation studies by removing key components: the self-attention and cross-attention mechanisms. These components are crucial for modeling relationships between keypoints and extracting structure-aware features. As shown in Table \ref{tab:ablation2}, without them, the network struggles to capture intra- and inter-image structure, resulting in degraded correspondence estimation. Furthermore, to validate the effectiveness of directly estimating 3D keypoints, we replaced this process with a relative pose regression. We evaluated two variations: averaging point-wise pose regression and directly regressing the relative pose with global features. As shown in Table \ref{tab:ablation2}, directly regressing 3D keypoint coordinates is significantly more effective than regressing the entire rotation matrix. By focusing on keypoint coordinate regression, our method captures the underlying structural relationships between different parts of the object, ultimately resulting in more reliable 3D correspondence and precise pose estimation.

\textbf{Effects of Mask Loss and Confidence Score.}  As shown in Table \ref{tab:ablation3}, both mask loss and confidence score estimation play important roles in improving performance. The mask loss helps the model focus on extracting foreground features, while the confidence score measures the reliability of the estimated 3D correspondences. 

\textbf{Effects of keypoint numbers.} In Table \ref{tab:ablation4}, we show the impact of the number of keypoints $N_{\text{kpt}}$. It can be observed that as $N_{\text{kpt}}$ increases, the performance improves, which is attributed to the fact that more keypoints help in better modeling the structure of the object. For balancing performance gains against computational efficiency (measured in MACs), we select $N_{\text{kpt}} = 48$ by default.

\begin{figure}[t]
    \centering
    \includegraphics[width=1\linewidth]{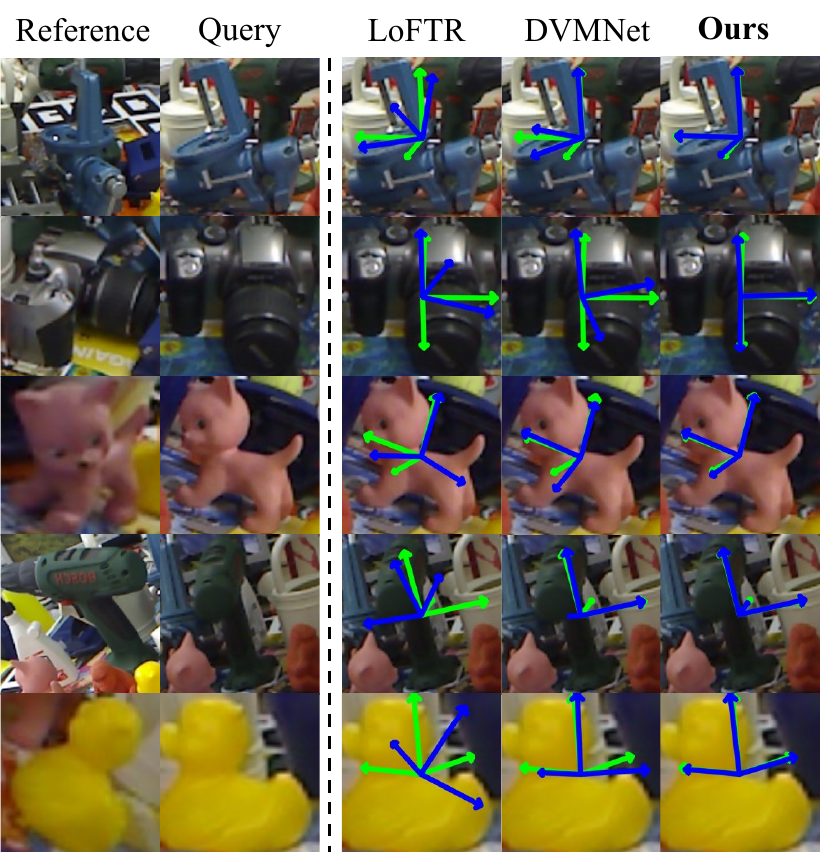}
        \caption{\textbf{Qualitative comparisons among LoFTR, DVMNet and our method.} We visualize the ground truth and predicted arrows. Blue indicates ground truth and green indicates prediction.}
    \label{fig:visual}
    \vspace{-12.5pt}
\end{figure}

\subsection{Visualization}
\textbf{Qualitative Results.} The qualitative results of LoFTR~\cite{sun2021loftr}, DVMNet~\cite{zhao2024dvmnet} and our method on the LineMOD dataset are shown in Figure \ref{fig:visual}. Specifically, we visualize the object pose depicted in the query image. The query object pose is determined as $\mathbf{R}_q = (\Delta \mathbf{R})^{-1} \mathbf{R}_r$, where $\mathbf{R}_r$ denotes the object pose in the reference image. The green and blue arrows represent the visualization of $\mathbf{R}_q$ calculated from the prediction and the ground truth, respectively. It can be observed that our method demonstrates more accurate performance compared to previous matching-based methods, especially in cases where there are larger viewpoint differences between the query and reference images.

\section{Conclusion}
\label{sec:conclusion}
In this paper, we propose a Structure-Aware Correspondence Learning method for Relative Pose Estimation. 
Specifically, we introduce a structure-aware keypoint extraction module to identify keypoints that can represent the structure of objects with different shapes and appearance. Furthermore, we propose a correspondence estimation module that models relationships between keypoints to extract structure-aware features, enabling robust 3D correspondence regression without explicit feature matching. Comprehensive experiments on three datasets demonstrate the effectiveness of our method.
\section{Acknowledgements}
This work was supported by the National Defense Science and Technology Foundation Strengthening Program Funding (Grant 2023-JCJQ-JJ-0219).
{
    \small
    \bibliographystyle{ieeenat_fullname}
    \bibliography{main}

@String(CVPR= {IEEE Conf. Comput. Vis. Pattern Recog.})

@String(ECCV= {Eur. Conf. Comput. Vis.})

@String(CVPR  = {CVPR})

@String(ECCV  = {ECCV})

@inproceedings{mousavian20196,
  title={6-dof graspnet: Variational grasp generation for object manipulation},
  author={Mousavian, Arsalan and Eppner, Clemens and Fox, Dieter},
  booktitle={Proceedings of the IEEE/CVF international conference on computer vision},
  pages={2901--2910},
  year={2019}
}

@article{wen2022you,
  title={You only demonstrate once: Category-level manipulation from single visual demonstration},
  author={Wen, Bowen and Lian, Wenzhao and Bekris, Kostas and Schaal, Stefan},
  journal={arXiv preprint arXiv:2201.12716},
  year={2022}
}

@article{wu2020grasp,
  title={Grasp proposal networks: An end-to-end solution for visual learning of robotic grasps},
  author={Wu, Chaozheng and Chen, Jian and Cao, Qiaoyu and Zhang, Jianchi and Tai, Yunxin and Sun, Lin and Jia, Kui},
  journal={Advances in Neural Information Processing Systems},
  volume={33},
  pages={13174--13184},
  year={2020}
}

@article{azuma1997survey,
  title={A Survey of Augmented Reality},
  author={Azuma, Ronald T},
  journal={Presence: Teleoperators and Virtual Environments/MIT press},
  year={1997}
}

@article{marchand2015pose,
  title={Pose estimation for augmented reality: a hands-on survey},
  author={Marchand, Eric and Uchiyama, Hideaki and Spindler, Fabien},
  journal={IEEE transactions on visualization and computer graphics},
  volume={22},
  number={12},
  pages={2633--2651},
  year={2015},
  publisher={IEEE}
}

@inproceedings{chen2017multi,
  title={Multi-view 3d object detection network for autonomous driving},
  author={Chen, Xiaozhi and Ma, Huimin and Wan, Ji and Li, Bo and Xia, Tian},
  booktitle={Proceedings of the IEEE conference on Computer Vision and Pattern Recognition},
  pages={1907--1915},
  year={2017}
}

@inproceedings{geiger2012we,
  title={Are we ready for autonomous driving? the kitti vision benchmark suite},
  author={Geiger, Andreas and Lenz, Philip and Urtasun, Raquel},
  booktitle={2012 IEEE conference on computer vision and pattern recognition},
  pages={3354--3361},
  year={2012},
  organization={IEEE}
}

@article{xiang2017posecnn,
  title={Posecnn: A convolutional neural network for 6d object pose estimation in cluttered scenes},
  author={Xiang, Yu and Schmidt, Tanner and Narayanan, Venkatraman and Fox, Dieter},
  journal={arXiv preprint arXiv:1711.00199},
  year={2017}
}

@inproceedings{wang2019densefusion,
  title={Densefusion: 6d object pose estimation by iterative dense fusion},
  author={Wang, Chen and Xu, Danfei and Zhu, Yuke and Mart{\'\i}n-Mart{\'\i}n, Roberto and Lu, Cewu and Fei-Fei, Li and Savarese, Silvio},
  booktitle={Proceedings of the IEEE/CVF conference on computer vision and pattern recognition},
  pages={3343--3352},
  year={2019}
}

@inproceedings{he2020pvn3d,
  title={Pvn3d: A deep point-wise 3d keypoints voting network for 6dof pose estimation},
  author={He, Yisheng and Sun, Wei and Huang, Haibin and Liu, Jianran and Fan, Haoqiang and Sun, Jian},
  booktitle={Proceedings of the IEEE/CVF conference on computer vision and pattern recognition},
  pages={11632--11641},
  year={2020}
}

@inproceedings{he2021ffb6d,
  title={Ffb6d: A full flow bidirectional fusion network for 6d pose estimation},
  author={He, Yisheng and Huang, Haibin and Fan, Haoqiang and Chen, Qifeng and Sun, Jian},
  booktitle={Proceedings of the IEEE/CVF conference on computer vision and pattern recognition},
  pages={3003--3013},
  year={2021}
}

@inproceedings{rad2017bb8,
  title={Bb8: A scalable, accurate, robust to partial occlusion method for predicting the 3d poses of challenging objects without using depth},
  author={Rad, Mahdi and Lepetit, Vincent},
  booktitle={Proceedings of the IEEE international conference on computer vision},
  pages={3828--3836},
  year={2017}
}

@article{fischler1981random,
  title={Random sample consensus: a paradigm for model fitting with applications to image analysis and automated cartography},
  author={Fischler, Martin A and Bolles, Robert C},
  journal={Communications of the ACM},
  volume={24},
  number={6},
  pages={381--395},
  year={1981},
  publisher={ACM New York, NY, USA}
}

@inproceedings{li2019cdpn,
  title={Cdpn: Coordinates-based disentangled pose network for real-time rgb-based 6-dof object pose estimation},
  author={Li, Zhigang and Wang, Gu and Ji, Xiangyang},
  booktitle={Proceedings of the IEEE/CVF international conference on computer vision},
  pages={7678--7687},
  year={2019}
}

@inproceedings{zakharov2019dpod,
  title={Dpod: 6d pose object detector and refiner},
  author={Zakharov, Sergey and Shugurov, Ivan and Ilic, Slobodan},
  booktitle={Proceedings of the IEEE/CVF international conference on computer vision},
  pages={1941--1950},
  year={2019}
}

@inproceedings{sundermeyer2018implicit,
  title={Implicit 3d orientation learning for 6d object detection from rgb images},
  author={Sundermeyer, Martin and Marton, Zoltan-Csaba and Durner, Maximilian and Brucker, Manuel and Triebel, Rudolph},
  booktitle={Proceedings of the european conference on computer vision (ECCV)},
  pages={699--715},
  year={2018}
}

@inproceedings{gao20206d,
  title={6d object pose regression via supervised learning on point clouds},
  author={Gao, Ge and Lauri, Mikko and Wang, Yulong and Hu, Xiaolin and Zhang, Jianwei and Frintrop, Simone},
  booktitle={2020 IEEE International Conference on Robotics and Automation (ICRA)},
  pages={3643--3649},
  year={2020},
  organization={IEEE}
}

@inproceedings{lin20216d,
  title={6D object pose estimation with pairwise compatible geometric features},
  author={Lin, Muyuan and Murali, Varun and Karaman, Sertac},
  booktitle={2021 IEEE International Conference on Robotics and Automation (ICRA)},
  pages={10966--10973},
  year={2021},
  organization={IEEE}
}

@inproceedings{liu2021kdfnet,
  title={Kdfnet: Learning keypoint distance field for 6d object pose estimation},
  author={Liu, Xingyu and Iwase, Shun and Kitani, Kris M},
  booktitle={2021 IEEE/RSJ International Conference on Intelligent Robots and Systems (IROS)},
  pages={4631--4638},
  year={2021},
  organization={IEEE}
}

@inproceedings{tian2020robust,
  title={Robust 6d object pose estimation by learning rgb-d features},
  author={Tian, Meng and Pan, Liang and Ang, Marcelo H and Lee, Gim Hee},
  booktitle={2020 IEEE International Conference on Robotics and Automation (ICRA)},
  pages={6218--6224},
  year={2020},
  organization={IEEE}
}

@inproceedings{li2022dcl,
  title={Dcl-net: Deep correspondence learning network for 6d pose estimation},
  author={Li, Hongyang and Lin, Jiehong and Jia, Kui},
  booktitle={European Conference on Computer Vision},
  pages={369--385},
  year={2022},
  organization={Springer}
}

@inproceedings{wang2019normalized,
  title={Normalized object coordinate space for category-level 6d object pose and size estimation},
  author={Wang, He and Sridhar, Srinath and Huang, Jingwei and Valentin, Julien and Song, Shuran and Guibas, Leonidas J},
  booktitle={Proceedings of the IEEE/CVF Conference on Computer Vision and Pattern Recognition},
  pages={2642--2651},
  year={2019}
}

@inproceedings{tian2020shape,
  title={Shape prior deformation for categorical 6d object pose and size estimation},
  author={Tian, Meng and Ang, Marcelo H and Lee, Gim Hee},
  booktitle={Computer Vision--ECCV 2020: 16th European Conference, Glasgow, UK, August 23--28, 2020, Proceedings, Part XXI 16},
  pages={530--546},
  year={2020},
  organization={Springer}
}

@inproceedings{lin2022category,
  title={Category-level 6d object pose and size estimation using self-supervised deep prior deformation networks},
  author={Lin, Jiehong and Wei, Zewei and Ding, Changxing and Jia, Kui},
  booktitle={European Conference on Computer Vision},
  pages={19--34},
  year={2022},
  organization={Springer}
}

@inproceedings{lin2023vi,
  title={Vi-net: Boosting category-level 6d object pose estimation via learning decoupled rotations on the spherical representations},
  author={Lin, Jiehong and Wei, Zewei and Zhang, Yabin and Jia, Kui},
  booktitle={Proceedings of the IEEE/CVF International Conference on Computer Vision},
  pages={14001--14011},
  year={2023}
}

@inproceedings{lin2024instance,
  title={Instance-adaptive and geometric-aware keypoint learning for category-level 6d object pose estimation},
  author={Lin, Xiao and Yang, Wenfei and Gao, Yuan and Zhang, Tianzhu},
  booktitle={Proceedings of the IEEE/CVF Conference on Computer Vision and Pattern Recognition},
  pages={21040--21049},
  year={2024}
}

@inproceedings{chen2024secondpose,
  title={Secondpose: Se (3)-consistent dual-stream feature fusion for category-level pose estimation},
  author={Chen, Yamei and Di, Yan and Zhai, Guangyao and Manhardt, Fabian and Zhang, Chenyangguang and Zhang, Ruida and Tombari, Federico and Navab, Nassir and Busam, Benjamin},
  booktitle={Proceedings of the IEEE/CVF Conference on Computer Vision and Pattern Recognition},
  pages={9959--9969},
  year={2024}
}

@inproceedings{zhang2022relpose,
  title={Relpose: Predicting probabilistic relative rotation for single objects in the wild},
  author={Zhang, Jason Y and Ramanan, Deva and Tulsiani, Shubham},
  booktitle={European Conference on Computer Vision},
  pages={592--611},
  year={2022},
  organization={Springer}
}

@inproceedings{lin2024relpose++,
  title={Relpose++: Recovering 6d poses from sparse-view observations},
  author={Lin, Amy and Zhang, Jason Y and Ramanan, Deva and Tulsiani, Shubham},
  booktitle={2024 International Conference on 3D Vision (3DV)},
  pages={106--115},
  year={2024},
  organization={IEEE}
}

@article{zhao20233d,
  title={3d-aware hypothesis \& verification for generalizable relative object pose estimation},
  author={Zhao, Chen and Zhang, Tong and Salzmann, Mathieu},
  journal={arXiv preprint arXiv:2310.03534},
  year={2023}
}

@inproceedings{besl1992method,
  title={Method for registration of 3-D shapes},
  author={Besl, Paul J and McKay, Neil D},
  booktitle={Sensor fusion IV: control paradigms and data structures},
  volume={1611},
  pages={586--606},
  year={1992},
  organization={Spie}
}

@inproceedings{zhao2024dvmnet,
  title={DVMNet: Computing Relative Pose for Unseen Objects Beyond Hypotheses},
  author={Zhao, Chen and Zhang, Tong and Dang, Zheng and Salzmann, Mathieu},
  booktitle={Proceedings of the IEEE/CVF Conference on Computer Vision and Pattern Recognition},
  pages={20485--20495},
  year={2024}
}

@inproceedings{nguyen2024nope,
  title={Nope: Novel object pose estimation from a single image},
  author={Nguyen, Van Nguyen and Groueix, Thibault and Ponimatkin, Georgy and Hu, Yinlin and Marlet, Renaud and Salzmann, Mathieu and Lepetit, Vincent},
  booktitle={Proceedings of the IEEE/CVF Conference on Computer Vision and Pattern Recognition},
  pages={17923--17932},
  year={2024}
}

@inproceedings{sun2021loftr,
  title={LoFTR: Detector-free local feature matching with transformers},
  author={Sun, Jiaming and Shen, Zehong and Wang, Yuang and Bao, Hujun and Zhou, Xiaowei},
  booktitle={Proceedings of the IEEE/CVF conference on computer vision and pattern recognition},
  pages={8922--8931},
  year={2021}
}

@inproceedings{sarlin2020superglue,
  title={Superglue: Learning feature matching with graph neural networks},
  author={Sarlin, Paul-Edouard and DeTone, Daniel and Malisiewicz, Tomasz and Rabinovich, Andrew},
  booktitle={Proceedings of the IEEE/CVF conference on computer vision and pattern recognition},
  pages={4938--4947},
  year={2020}
}

@article{lowe2004distinctive,
  title={Distinctive image features from scale-invariant keypoints},
  author={Lowe, David G},
  journal={International journal of computer vision},
  volume={60},
  pages={91--110},
  year={2004},
  publisher={Springer}
}

@inproceedings{reizenstein2021common,
  title={Common objects in 3d: Large-scale learning and evaluation of real-life 3d category reconstruction},
  author={Reizenstein, Jeremy and Shapovalov, Roman and Henzler, Philipp and Sbordone, Luca and Labatut, Patrick and Novotny, David},
  booktitle={Proceedings of the IEEE/CVF international conference on computer vision},
  pages={10901--10911},
  year={2021}
}

@inproceedings{hinterstoisser2013model,
  title={Model based training, detection and pose estimation of texture-less 3d objects in heavily cluttered scenes},
  author={Hinterstoisser, Stefan and Lepetit, Vincent and Ilic, Slobodan and Holzer, Stefan and Bradski, Gary and Konolige, Kurt and Navab, Nassir},
  booktitle={Computer Vision--ACCV 2012: 11th Asian Conference on Computer Vision, Daejeon, Korea, November 5-9, 2012, Revised Selected Papers, Part I 11},
  pages={548--562},
  year={2013},
  organization={Springer}
}

@inproceedings{deitke2023objaverse,
  title={Objaverse: A universe of annotated 3d objects},
  author={Deitke, Matt and Schwenk, Dustin and Salvador, Jordi and Weihs, Luca and Michel, Oscar and VanderBilt, Eli and Schmidt, Ludwig and Ehsani, Kiana and Kembhavi, Aniruddha and Farhadi, Ali},
  booktitle={Proceedings of the IEEE/CVF Conference on Computer Vision and Pattern Recognition},
  pages={13142--13153},
  year={2023}
}

@inproceedings{he2017mask,
  title={Mask r-cnn},
  author={He, Kaiming and Gkioxari, Georgia and Doll{\'a}r, Piotr and Girshick, Ross},
  booktitle={Proceedings of the IEEE international conference on computer vision},
  pages={2961--2969},
  year={2017}
}

@inproceedings{kirillov2023segment,
  title={Segment anything},
  author={Kirillov, Alexander and Mintun, Eric and Ravi, Nikhila and Mao, Hanzi and Rolland, Chloe and Gustafson, Laura and Xiao, Tete and Whitehead, Spencer and Berg, Alexander C and Lo, Wan-Yen and others},
  booktitle={Proceedings of the IEEE/CVF International Conference on Computer Vision},
  pages={4015--4026},
  year={2023}
}

@inproceedings{weinzaepfel2023croco,
  title={CroCo v2: Improved cross-view completion pre-training for stereo matching and optical flow},
  author={Weinzaepfel, Philippe and Lucas, Thomas and Leroy, Vincent and Cabon, Yohann and Arora, Vaibhav and Br{\'e}gier, Romain and Csurka, Gabriela and Antsfeld, Leonid and Chidlovskii, Boris and Revaud, J{\'e}r{\^o}me},
  booktitle={Proceedings of the IEEE/CVF International Conference on Computer Vision},
  pages={17969--17980},
  year={2023}
}

@article{arun1987least,
  title={Least-squares fitting of two 3-D point sets},
  author={Arun, K Somani and Huang, Thomas S and Blostein, Steven D},
  journal={IEEE Transactions on pattern analysis and machine intelligence},
  year={1987},
  volume={PAMI-9},
  number={5},
  pages={698-700},
  publisher={IEEE}
}

@inproceedings{zhou2019continuity,
  title={On the continuity of rotation representations in neural networks},
  author={Zhou, Yi and Barnes, Connelly and Lu, Jingwan and Yang, Jimei and Li, Hao},
  booktitle={Proceedings of the IEEE/CVF conference on computer vision and pattern recognition},
  pages={5745--5753},
  year={2019}
}

@article{su2024roformer,
  title={Roformer: Enhanced transformer with rotary position embedding},
  author={Su, Jianlin and Ahmed, Murtadha and Lu, Yu and Pan, Shengfeng and Bo, Wen and Liu, Yunfeng},
  journal={Neurocomputing},
  volume={568},
  pages={127063},
  year={2024},
  publisher={Elsevier}
}

@article{kingma2014adam,
  title={Adam: A method for stochastic optimization},
  author={Kingma, Diederik P},
  journal={arXiv preprint arXiv:1412.6980},
  year={2014}
}

@inproceedings{goodwin2022zero,
  title={Zero-shot category-level object pose estimation},
  author={Goodwin, Walter and Vaze, Sagar and Havoutis, Ioannis and Posner, Ingmar},
  booktitle={European Conference on Computer Vision},
  pages={516--532},
  year={2022},
  organization={Springer}
}

@article{simonyan2014very,
  title={Very deep convolutional networks for large-scale image recognition},
  author={Simonyan, Karen and Zisserman, Andrew},
  journal={arXiv preprint arXiv:1409.1556},
  year={2014}
}

@article{vaswani2017attention,
  title={Attention is all you need},
  author={Vaswani, A},
  journal={Advances in Neural Information Processing Systems},
  year={2017}
}

@inproceedings{iclr2025spherepose,
    title={Learning Shape-Independent Transformation via Spherical Representations for Category-Level Object Pose Estimation},
    author={Ren, Huan and Yang, Wenfei and Liu, Xiang and Zhang, Shifeng and Zhang, Tianzhu},
    booktitle={The Thirteenth International Conference on Learning Representations},
    year={2025}
}

@inproceedings{cvpr2025spotpose,
    title={Rethinking Correspondence-based Category-Level Object Pose Estimation},
    author={Ren, Huan and Yang, Wenfei and Zhang, Shifeng and Zhang, Tianzhu},
    booktitle={Proceedings of the IEEE/CVF Conference on Computer Vision and Pattern Recognition (CVPR)},
    year={2025}
}
}


\end{document}